\documentclass{isprs} %
\usepackage{setspace}
\usepackage{geometry} %
\usepackage{epstopdf}
\usepackage[labelsep=period]{caption}  %
\usepackage[british]{babel} 
\usepackage[hang]{footmisc}
\usepackage{algorithm}
\usepackage{algpseudocode}
\usepackage{amsmath}
\usepackage{amssymb}
\usepackage{graphicx}
\usepackage{subcaption}
\usepackage{url}
\usepackage{multirow}
\usepackage{float}

\begin{document}

\title{GeoSceneGraph: Geometric Scene Graph Diffusion Model for Text-guided 3D Indoor Scene Synthesis}
\date{}

\author{Antonio Ruiz \textsuperscript{1,2}, Tao Wu \textsuperscript{1}, Andrew Melnik\textsuperscript{2}, Qing Cheng \textsuperscript{1}, Xuqin Wang \textsuperscript{1}, Lu Liu \textsuperscript{1}, Yongliang Wang \textsuperscript{1}, Yanfeng Zhang \textsuperscript{1}, Helge Ritter \textsuperscript{2}}

\address{\textsuperscript{1}Huawei Technologies \\ \textsuperscript{2}Center for Cognitive Interaction Technology (CITEC), Bielefeld University}

\abstract{Methods that synthesize indoor 3D scenes from text prompts have wide-ranging applications in film production, interior design, video games, virtual reality, and synthetic data generation for training embodied agents. Existing approaches typically either train generative models from scratch or leverage vision-language models (VLMs). While VLMs achieve strong performance, particularly for complex or open-ended prompts, smaller task-specific models remain necessary for deployment on resource-constrained devices such as extended reality (XR) glasses or mobile phones. However, many generative approaches that train from scratch overlook the inherent graph structure of indoor scenes, which can limit scene coherence and realism. Conversely, methods that incorporate scene graphs either demand a user-provided semantic graph, which is generally inconvenient and restrictive, or rely on ground-truth relationship annotations, limiting their capacity to capture more varied object interactions. To address these challenges, we introduce GeoSceneGraph, a method that synthesizes 3D scenes from text prompts by leveraging the graph structure and geometric symmetries of 3D scenes, without relying on predefined relationship classes. Despite not using ground-truth relationships, GeoSceneGraph achieves performance comparable to methods that do. Our model is built on equivariant graph neural networks (EGNNs), but existing EGNN approaches are typically limited to low-dimensional conditioning and are not designed to handle complex modalities such as text. We propose a simple and effective strategy for conditioning EGNNs on text features, and we validate our design through ablation studies.
}


\maketitle

\section{Introduction}

\label{sec:Introduction}

The automatic synthesis of high-quality 3D indoor scenes is a problem that presents significant challenges in computer vision and computer graphics. Key challenges that need to be addressed include: 1) generating realistic and diverse 3D indoor scenes, where the aesthetics of objects are visually appealing, and their arrangement is meaningful, reflecting real-world functionality and spatial organization; and 2) ensuring the model can process control signals to meet user-specific requirements. This problem is highly relevant, as automating scene generation can significantly reduce the time and cost of producing computer-generated imagery (CGI) for applications such as film production, interior design, video games, and extended reality (XR), a term that encompasses augmented reality (AR), mixed reality (MR), and virtual reality (VR). Moreover, synthetic 3D scenes can provide rich ground-truth data for training models in 3D scene understanding and reconstruction, and can also serve as environments for training embodied agents.

Recently, this problem has been addressed by deep generative methods that employ variational autoencoders (VAEs) \cite{kingma2013auto}, generative adversarial networks (GANs) \cite{goodfellow2020generative}, autoregressive models or diffusion models \cite{sohl2015deep,ho2020denoising}. Approaches based on GANs \cite{zhang2020deep,yang2021indoor} are able to synthesize 3D scenes, but they often struggle with scene diversity, and the generation process is difficult to control. VAE-based methods  \cite{purkait2020sg,yang2021scene} offer better diversity in the generated samples, but the overall quality tends to be lower. On the other hand, auto-regressive models \cite{paschalidou2021atiss,nie2023learning,wang2021sceneformer} fail to capture semantic relationships between objects, and and they are prone to accumulating prediction errors. Although some diffusion-based methods \cite{tang2024diffuscene,hu2024mixed,maillard2025debara} tackle these issues, they do not incorporate the graph structure of the 3D scene, which adversely impacts their performance. By contrast, other diffusion-based approaches \cite{zhai2023commonscenes,zhai2024echoscene,lin2024instructscene} show that accounting for the inherent graph structure of the scene enhances both quality and controllability. Nonetheless, methods relying on the scene graph as a direct control signal \cite{zhai2023commonscenes,zhai2024echoscene} tend to be less user-friendly. InstructScene \cite{lin2024instructscene} overcomes these challenges by allowing text control signals and generating semantic scene graphs with a two-stage approach. Nevertheless, their semantic scene graph generation (first stage) relies on a static and predefined vocabulary of semantic relations for the graph edges, limiting their ability to be train with open-vocabulary relationships. Previous methods have also overlooked the inherent geometric symmetries of 3D scenes, such as translations and rotations, referred to as the special Euclidean group in 3 dimensions, SE(3). However, recent works on 3D  molecule generation \cite{hoogeboom2022equivariant,xu2023geometric} demonstrate that leveraging these symmetries can significantly enhance performance and improve generalization.

Vision-language models \cite{feng2023layoutgpt,ccelen2024design,yang2024holodeck,sun2025layoutvlm,yang2025sceneweaver} have also been applied to solve this problem with strong success, particularly for open-ended prompts. However, their main limitation is the high computational cost associated with inference, which restricts their deployment on resource-constrained or offline devices. This makes it still important to develop smaller models capable of solving the task efficiently. For the rest of the paper, we will not further discuss VLM-based approaches and instead concentrate on task-specific methods, since our method belongs to this class.

To address shortcomings of previous task-specific methods, we propose GeoSceneGraph, a novel diffusion-based approach that incorporates the inherent graph structure and geometric symmetries of 3D scenes without relying on ground-truth object relations from a static, predefined vocabulary. In addition, we introduce a text-conditioning strategy for EGNNs. This strategy first fuses input feature vectors with the diffusion time-step condition through a ResNet \cite{he2016deep} and self-attention Transformer\cite{vaswani2017attention} blocks, then integrates the text context into the message passing steps of the EGNN.

Our main contributions can be summarized as follows:

\begin{itemize}

    \item We introduce a method for text-guided 3D indoor scene synthesis based on equivariant diffusion models \cite{hoogeboom2022equivariant}. Our approach leverages the scene graph structure and its geometric symmetries without relying on ground-truth semantic edges. Despite not relying on ground-truth edges, our method still achieves competitive performance compared to approaches that use them.
    
    \item We propose a novel conditioning mechanism for equivariant diffusion models, which is able to handle complex control signals such as text.

    \item We present various ablation studies exploring different conditioning strategies that validate our design choices.
     
\end{itemize}

\section{Method}
\label{sec:Method}

\begin{figure*}[ht!]
  \centering
  \includegraphics[width=0.90\textwidth]{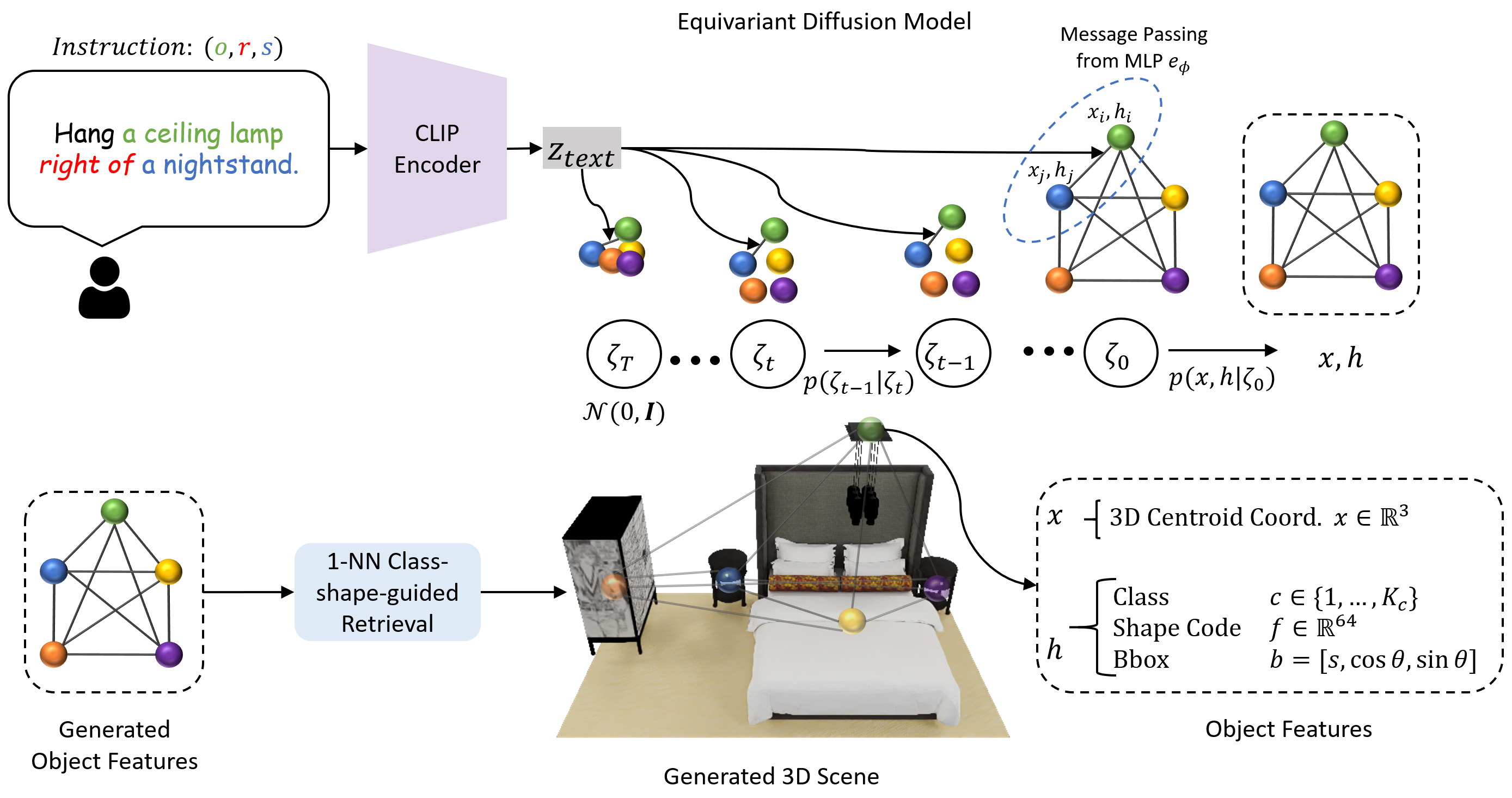}
  \caption{Overall pipeline of our method. First, we encode the text prompt with a CLIP encoder, and use this text embedding to condition the diffusion process through all message passing steps in the denoising EGNN. Once object features are sampled with the denoising process, we generate the scene by retrieving objects via 1-NN search and positioning them with 3D coordinates and bounding box parameters.
}
  \label{fig:pipeline}
\end{figure*}

\begin{figure}[ht!]
  \centering
  \includegraphics[width=0.5\textwidth]{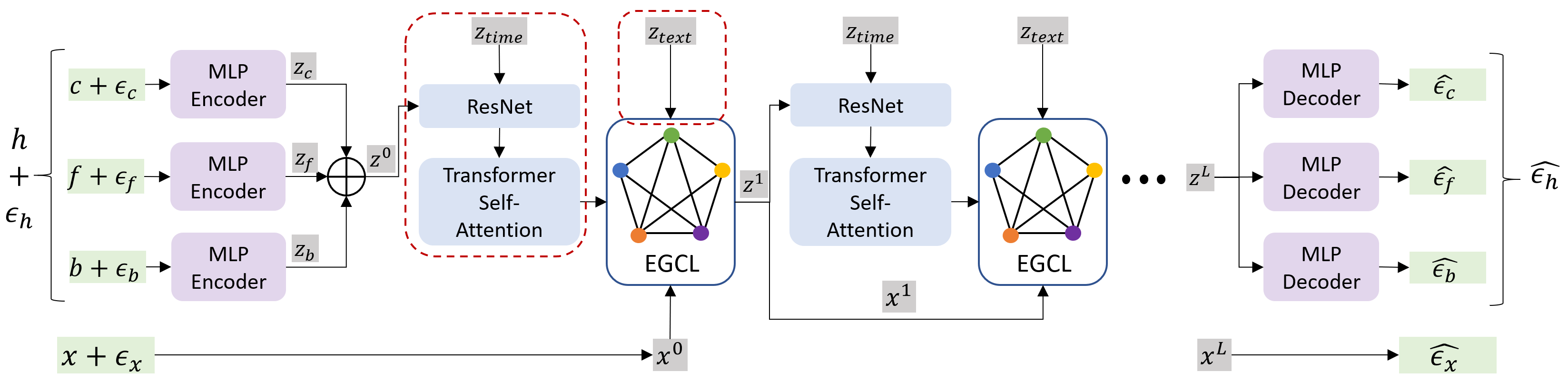}
  \caption{EGNN Architecture. Our EGNN architecture denoises the noisy node features \([x + \epsilon_x,\, h + \epsilon_h]\) through a three-phase pipeline composed of MLP encoders, EGCL layers and MLP decoders. Our novelty is highlighted by the dotted red boxes.}
  \label{fig:egnn-architecture}
\end{figure}

In this section we describe our method. Following previous works \cite{paschalidou2021atiss,tang2024diffuscene,lin2024instructscene}, we denote a collection of 3D scenes with the set \(\mathcal{S} := \{\mathcal{S}_1, \mathcal{S}_2, \ldots,  \mathcal{S}_{M}\} \). Then, we define a 3D scene \(\mathcal{S}_k\) as a collection of 3D objects, \(\mathcal{S}_k := \{o_{k1}, o_{k2}, \ldots, o_{kN_k}\} \) arranged according to a specific spatial layout relative to a world coordinate system, where the origin is placed at the center of the floor. Each object \(o_{ki}\) is associated with layout and appearance features. The layout features include \(x_{ki} \in \mathbb{R}^3\), representing the position of the centroid of the object in the world coordinate system,
\(\theta_{ki} \in [0, 2\pi]\) (or its unit circle representation \((\cos \theta_{ki}, \sin \theta_{ki})\)) indicates the yaw-axis rotation angle, restricting the object rotation to the vertical axis, and the size of its bounding box \(s_{ki} \in \mathbb{R}^3\). The appearance features consist of the object class \(c_{ki} \in \{1, \ldots, K_c\}\), where \(K_c\) is the number of object classes in the dataset, and shape code \(f_{ki} \in \mathbb{R}^d\). The shape codes are extracted from OpenCLIP \texttt{ViT-bigG/14} embeddings \cite{liu2023openshape}, and their dimensionality is reduced with a VAE (see more details in Section \ref{sec:shape-auto-encoder}). Given these features, a 3D scene \(\mathcal{S}_k\), can be constructed by retrieving objects from a database with prebuilt 3D object models using a two-step filtering process. First, the selection is made based on the class \(c_{ki}\), and then, on the nearest neighbor of the shape feature \(f_{ki}\). Once the objects are chosen, the respective layout features are applied, i.e., object \(o_{ki}\) is translated to position \(x_{ki}\), rotated by \(\theta_{ki}\) about the vertical axis, and resized according to \(s_{ki}\). We refer to Figure \ref{fig:pipeline} to see the pipeline of our method.

Given a collection of 3D scenes \(\mathcal{S}\), we can learn a parameterized generative model \(p_{\phi} (\mathcal{S}) \) that generates both appearance and layout features, and then use the process described above to synthesize a novel 3D scene. Furthermore, to incorporate text control signals, we can also learn a conditional generative model \(p_{\phi} (\mathcal{S}|y) \), where \(y\) denotes the user text prompt.

For clarity in notation, we will omit the subscript \(ki\) from here onwards. Inspired by equivariant diffusion models (EDM) \cite{hoogeboom2022equivariant,xu2023geometric} used for molecular generation, we model a geometric scene graph \(G=(V,E)\), corresponding to scene \(S\), similar to a molecular graph, where the nodes \(v \in V\) are endowed with 3D geometric coordinates \(x\) and node features \(h\), i.e. \(v = \langle x, h \rangle\). In our scenario, \(x = \{x_1, x_2, \ldots, x_{N_i}\}\) represents the 3D coordinates of the centroids of the objects in the 3D scene, while the rest of node features are denoted by \(h = \{h_1, h_2, \ldots, h_{N_i}\}\). Each feature vector \(h_{i}\) is composed of three components: \(h_{i} = [c_{i}, f_{i}, b_{i}]\), where \(c_{i}\) is the class, \(f_{i}\) is the shape code, and \(b_{i} = [s_{i}, \cos \theta_{i}, \\
\sin \theta_{i}]\) represents the scaled and rotated bounding box of object \(o_{i}\). Here, \([\cdot, \cdot]\) denotes concatenation of the inputs. For the edges \(E\), we assume full connectivity between the nodes. The goal is to train an EDM \(p_\phi\) conditioned on text prompts to learn the denoising process of the node features and 3D coordinates from the geometric scene graphs \(G = (V, E)\). This will allows us to generate layout and appearance features of the objects to construct a novel 3D scene. We extend the EGNN framework from the original EDM paper \cite{hoogeboom2022equivariant} (Equation 12), adapting it to handle higher-dimensional feature vectors and more complex conditioning, such as text control signals. We review the concepts of equivariance and invariance in Section~\ref{sec:equivariance-and-invariance}, and provide a detailed overview of equivariant graph neural networks (EGNNs) in Section~\ref{sec:EGNN}.

\subsection{Text-aligned Shape Autoencoder}
\label{sec:shape-auto-encoder}
Similar to InstructScene \cite{lin2024instructscene}, we obtain the object shape codes using OpenCLIP \texttt{ViT-bigG/14} \cite{liu2023openshape}, which is a model that aligns CLIP \cite{radford2021learning} embeddings with 3D objects. The advantage of using OpenCLIP embeddings is that they ease the learning for the model to capture more nuanced and complex object descriptions, a capability that is more challenging to learn when the shape codes are extracted solely from point clouds, 
as in DiffuScene \cite{tang2024diffuscene}. However, instead of employing a vector-quantized variational autoencoder (VQ-VAE) \cite{van2017neural} as is done in InstructScene, we use a standard VAE \cite{kingma2013auto} to reduce the high dimensionality of the OpenCLIP embeddings. Our method works well with embeddings in a continuous latent space, which allows us to use a lower dimensionality of 64 instead of 256 (as in InstructScene). The original dimensionality of OpenCLIP embeddings is 1280.

\subsection{Equivariance and Invariance}
\label{sec:equivariance-and-invariance}
Let  \(\rho(\mathfrak{g})\) be the linear representation related to the group element \(\mathfrak{g} \in \mathfrak{G}\), where \(\mathfrak{G}\) is a group. A function \(f\) is \(\mathfrak{G}\)\textit{-equivariant} to the action of a group 
\(\mathfrak{G}\) if:
 \(f(\rho(\mathfrak{g})x) = \rho(\mathfrak{g})f(x), \
\text{for all } \mathfrak{g} \in \mathfrak{G}\),
i.e., the group action has the same effect if it is applied first to the input \(x\) or after to the output \(f(x)\)
\cite{bronstein2021geometric}. In the case of invariance, applying the group action on the input \(x\) has no effect on the output \(f(x)\). In this work, we make use of the special Euclidean group \(SE(3)\), that is generated by orthogonal rotation matrices \(R \in SO(3)\) and translation vectors \(t \in \mathbb{R}^3\). Let \( x = (x_1, x_2, \ldots, x_N) \in \mathbb{R}^{N \times 3} \) denote a point cloud with \( N \) points, and let \( h = (h_1, h_2, \ldots, h_N) \in \mathbb{R}^{N \times n_f} \) represent the concatenated features of these points, where \( n_f \) is the dimension of each feature vector \( h_i \). Then, we want a vector-valued function \((z_x, z_h) = f(x, h)\) that is \(SE(3)\)-invariant on \(h\) and \(SE(3)\)-equivariant on \(x\), i.e.:
\begin{equation}
\label{invariant-equivariant}
(Rz_x+t, z_h) = f(Rx+t, h).    
\end{equation}

\subsection{Equivariant Graph Neural Networks (EGNNs)}
\label{sec:EGNN}
\(E(n)\)-equivariant graph neural networks \cite{satorras2021n} satisfy the constraint from Eq. \ref{invariant-equivariant}. They have been successfully used for molecular generation \cite{hoogeboom2022equivariant,xu2023geometric}. In these works, all pairwise interactions between atoms are taken into account, and therefore all edges \(e_{ij} \in E,\ i \neq j\), are included in the EGNN. An EGNN is composed of multiple equivariant graph convolutional layers (EGCLs), where \([x^{l+1},\,h^{l+1}] = \mathrm{EGCL}\bigl[x^{l},h^{l}\bigr]\) are defined as follows \cite{hoogeboom2022equivariant}:
\begin{multline}
\label{eq:egcl}
m_{ij}^l = e_\phi\bigl(h_{i}^{l}, h_{j}^{l}, d_{ij}^{2}, a_{ij}\bigr),\ \
h_{i}^{l+1} = h_\phi\!\Bigl(
  h_{i}^{l},\, \sum_{j \neq i} m_{ij}^l \tilde{e}_{ij}^l
\Bigr), \\
x_{i}^{l+1} = x_{i}^{l}
+ \sum_{j \neq i}\,\frac{x_{j}^{l} - x_{i}^{l}}{d_{ij} + 1}\,
x_\phi\bigl(h_{i}^{l}, h_{j}^{l}, d_{ij}^2, a_{ij}\bigr),
\end{multline}%
where \(l\) indexes the EGCL layer, \(d_{ij} = \left\|x_i^l - x_j^l\right\|_2 \) is the Euclidean distance between the the coordinates \(x_i^l\) and \(x_j^l\) at layer \(l\), \(a_{ij}\) are optional edge attributes, and \(\tilde{e}_{ij}^l = m_\phi(m_{ij}^l)\). The functions  \(e_\phi, h_\phi, m_\phi \ \text{and} \  x_\phi\) are parameterized with learnable multi-layer perceptrons (MLPs). An EGNN is composed of a set of \(L\) EGCL layers, where the input of the first layer is \([x^0, h^0]\). We use this framework and extend it to our problem of 3D scene generation (see Section \ref{sec:EDM-for-3D-scene-graph}).

\subsection{EDM for 3D Scene Graph}
\label{sec:EDM-for-3D-scene-graph}
In this section we present an extension of the EGNN framework, which we refer to as 3D Scene Equivariant Graph Neural Network (3DS-EGNN). Our approach takes as input both the 3D coordinates \(x\) of objects and their associated features \(h\), denoted as \(x^0\) and \(h^0\), respectively. The framework consists of three main phases (see Figure \ref{fig:egnn-architecture}): encoding, EGCL layers, and decoding.

For the encoding phase, following prior works \cite{paschalidou2021atiss,tang2024diffuscene,lin2024instructscene}, each feature type in \(h\), specifically, the class \(c\), shape code \(f\), and bounding box \(b\), is processed using a multi-layer perceptron (MLP). This produces latent representations: \(z_c = E_\phi^c(c), z_f = E_\phi^f(f), z_b = E_\phi^b(b).\)
These latent variables are then combined to form a joint representation: \(
z = E_\phi^h\bigl([z_c, z_f, z_b]\bigr)\), which captures all the input features from \(h\). To incorporate text prompt conditioning for the  3D scene generation, we employ a CLIP text encoder \cite{radford2021learning} to extract text embeddings: \(
z_{\text{text}} = E^{CLIP}(y)\).
Additionally, we encode the diffusion process timestep \(t\) using positional encodings \cite{vaswani2017attention}, followed by an encoder:
\(z_{\text{time}} = E_\phi^t(\text{PE}(t))\). All these embeddings, namely \(z\), \(z_{\text{text}}\), and \(z_{\text{time}}\), are then utilized within the EGCL layers.

For the second phase, we use \(L\) EGCL layers for the 3DS-EGNN. At each layer, indexed by \(l\), we combine \(z^l\) and \(z_{time}\) using a ResNet \cite{he2016deep} block \(RN_\phi^l\), followed by a Transformer \cite{vaswani2017attention} block \(T_\phi^l\). Then, we compute the scaled vector difference between the node coordinates as \(\delta_{ij}^l = \frac{x_{j}^{l} - x_{i}^{l}}{\left\|x_{j}^{l} - x_{i}^{l}\right\|_2 + 1}\), and use the text embedding \(z_{text}\) as edge attributes. Afterwards, we update the latent variables \(z^l\) and the node coordinates \(x^l\) using MLPs \(e_\phi^l, m_\phi^l, z_\phi^l, x_\phi^l\). After processing \(x^0\) and \(z^0\) through the \(L\) EGCL layers, we obtain \(x^L\) and \(z^L\).

Finally, in the third phase, we decode \(z^L\) back into the original input features \(h^L\) using three distinct decoders: \(\hat{c} = D_\phi^c(z^l)\), \(\hat{f} = D_\phi^f(z^l)\), and \(\hat{b} = D_\phi^b(z^l)\). The coordinates \(x^L\) remain unchanged. The complete 3DS-EGNN framework is presented in Algorithm \ref{alg:3DS-EGNN}.
The key differences between our 3DS-EGNN framework and the original EGNN from EDM are as follows. First, we use ResNet and Transformer blocks to inject the time step embedding, which seems to be more effective for handling high-dimensional latent vectors (see Section \ref{sec:ablation}). Then, we merge the text embedding at each message passing and coordinate update step. In the original framework, they were simply appended to the node features. Another difference is that we model the geometric interaction between nodes using the scaled vector difference between their coordinates as opposed to the Euclidean distance.

\begin{algorithm}
\caption{3DS-EGNN Framework}\label{alg:3DS-EGNN}
\begin{algorithmic}
\Require Node features \([x^0, h^0]\), time step \(t\), text prompt \(y\)
\State \hspace{0.8cm}Encoders \(E_\phi^h, E_\phi^c, E_\phi^f, E_\phi^b, E^{CLIP}, E_\phi^t \)
\State \hspace{0.8cm}Decoders \( D_\phi^c, D_\phi^f, D_\phi^b \)
\State \hspace{0.8cm}EGCL layers \(\{e_\phi^l, m_\phi^l, z_\phi^l, x_\phi^l, RN_\phi^l, T_\phi^l\}_{l=0}^{L-1} \)
\State Encode input features \(h^0 = [c,f,b]\), time \(t\) and text \(y\):
\State \hspace{0.8cm}\(z^0 = E_\phi^h\bigl([E_\phi^c(c), E_\phi^f(f), E_\phi^b(b)]\bigr)\)
\State \hspace{0.8cm}\(z_{text} = E^{CLIP}(y), \ \ z_{time} = E_\phi^t(PE(t)) \)
\For {\(l\) in \(0, \ldots, L-1\)}
\State Compute \(z^l = T_\phi^l( RN_\phi^l(z^l, z_{time}))\)
\State Compute \(m_{ij}^l = e_\phi^l\bigl(z_{i}^{l}, z_{j}^{l}, \delta_{ij}^l, z_{text}\bigr)\)
\State Update \(x^{l+1}_i\) and \(z^{l+1}_i\):
\State \(\ \ \ z_{i}^{l+1} = z_\phi^l\!\Bigl(
  z_{i}^{l},\, \sum_{j \neq i} m_{ij}^l \  m_\phi^l(m_{ij}^l)
\Bigr)\)
\State \( \ \ \ x_{i}^{l+1} = x_{i}^{l}
+ \sum_{j \neq i}\,\delta_{ij}^l\,
x_\phi^l\bigl(z_{i}^{l}, z_{j}^{l}, \delta_{ij}^l, z_{text}\bigr)\)
\EndFor
\State Decode \(z^L\) back to \(h^L = [\hat{c},\hat{f},\hat{b}]\):
\State \hspace{0.8cm}\(\hat{c} = D_\phi^c(z^L), \ \hat{f} = D_\phi^f(z^L), \ \hat{b} = D_\phi^b(z^L)\)
\State \Return \([x^L,h^L]\)
\end{algorithmic}
\end{algorithm}

The training and sampling procedures of the EDM remain largely as in the original EDM paper \cite{hoogeboom2022equivariant}, with the key difference that we employ the 3DS-EGNN outlined above instead of the original EGNN framework. We present the respective algorithms in Algorithm \ref{alg:SG-EDM-training} and Algorithm \ref{alg:SG-EDM-sampling}. We replace the variable \(z_t\) from the original EDM with \(\zeta_t\)  to avoid confusion with the latent variable resulting from the encoding phase in the 3DS-EGNN.

\begin{algorithm}
\caption{SG-EDM Training}\label{alg:SG-EDM-training}
\begin{algorithmic}
\Require Data point \(x,h = [c,f,b]\), text prompt \(y\)
\State \hspace{0.8cm}3DS-EGNN \(f_\phi\) 
\State Sample \(t \sim \mathcal{U}(1,\ldots T), \ \epsilon \sim \mathcal{N}(0, \mathbf{I}) \)

\State Substract the center of gravity in \(\epsilon_x\) in \(\epsilon = [\epsilon_x, \epsilon_h]\)
\State Compute \(\zeta_t = \sqrt{\bar{\alpha}_t}[x_0,h_0] + \sqrt{1-\bar{\alpha}_t}\epsilon \)
\State Minimize \(\left\|\epsilon - f_\phi(\zeta_t, t, y)\right\|_2^2 \)
\end{algorithmic}
\end{algorithm}

\begin{algorithm}
\caption{SG-EDM Sampling}\label{alg:SG-EDM-sampling}
\begin{algorithmic}
\Require Text prompt \(y\), trained 3DS-EGNN \(f_\phi\)
\State Sample \(\zeta_T \sim \mathcal{N}(0, \mathbf{I}) \)
\For {\(t\) in \(T, T-1, \ldots 1\)}
\State Sample \(\epsilon \in \mathcal{N}(0, \mathbf{I}) \)
\State Substract the center of gravity in \(\epsilon_x\) in \(\epsilon = [\epsilon_x, \epsilon_h]\)
\State Compute \(\zeta_{t-1} = \frac{1}{\sqrt{\alpha_t}}\bigl(\zeta_t - \frac{1-\alpha_t}{\sqrt{1-\bar{\alpha}_t}}f_\phi(\zeta_t,t,y)\bigr) + \sigma_t\epsilon \)
\EndFor
\State \Return \([x,h] \sim p(x,h \mid \zeta_0)\)
\end{algorithmic}
\end{algorithm}

\section{Experiments}
\label{sec:experiments}

\subsection{Experimental Settings}

\begin{figure*}[t]
\centering

\begin{subfigure}[t]{0.15\textwidth}
  \centering
  \includegraphics[width=\linewidth]{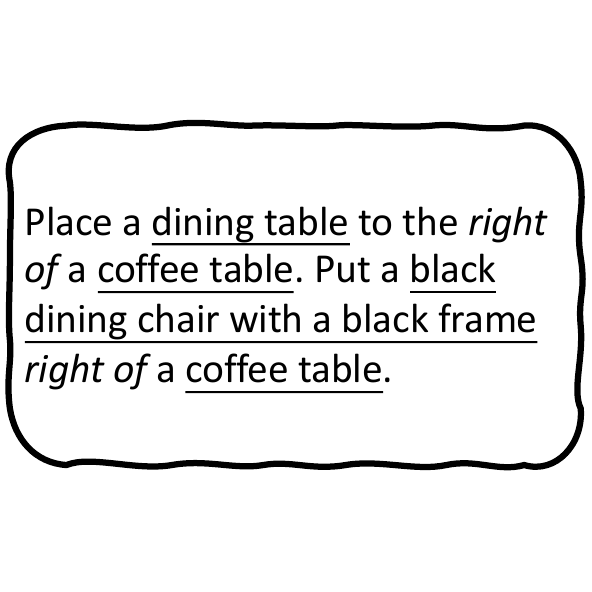}
\end{subfigure}\hfill
\begin{subfigure}[t]{0.15\textwidth}
  \centering
  \includegraphics[width=\linewidth]{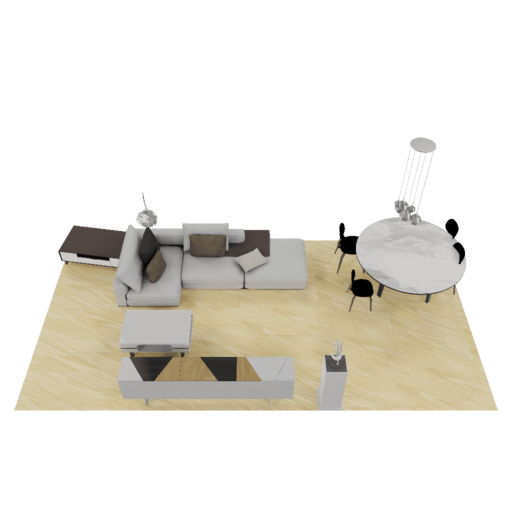}
\end{subfigure}\hfill
\begin{subfigure}[t]{0.15\textwidth}
  \centering
  \includegraphics[width=\linewidth]{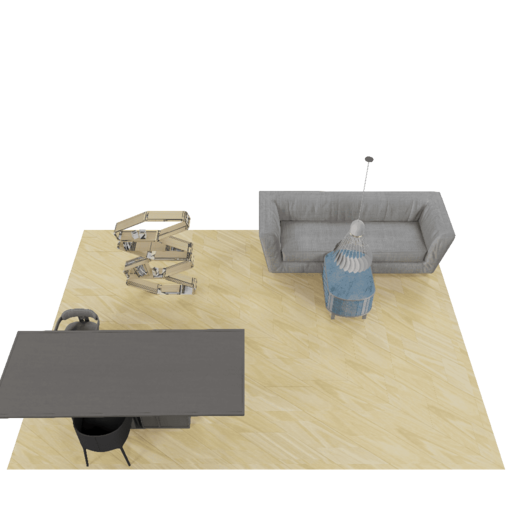}
\end{subfigure}\hfill
\begin{subfigure}[t]{0.15\textwidth}
  \centering
  \includegraphics[width=\linewidth]{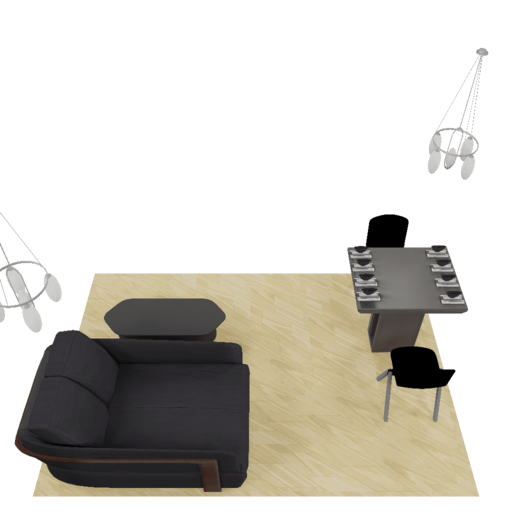}
\end{subfigure}\hfill
\begin{subfigure}[t]{0.15\textwidth}
  \centering
  \includegraphics[width=\linewidth]{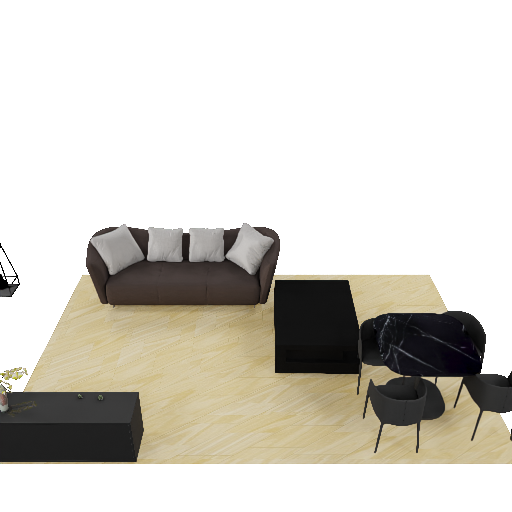}
\end{subfigure}

\par\medskip

\begin{subfigure}[t]{0.15\textwidth}
  \centering
  \includegraphics[width=\linewidth]{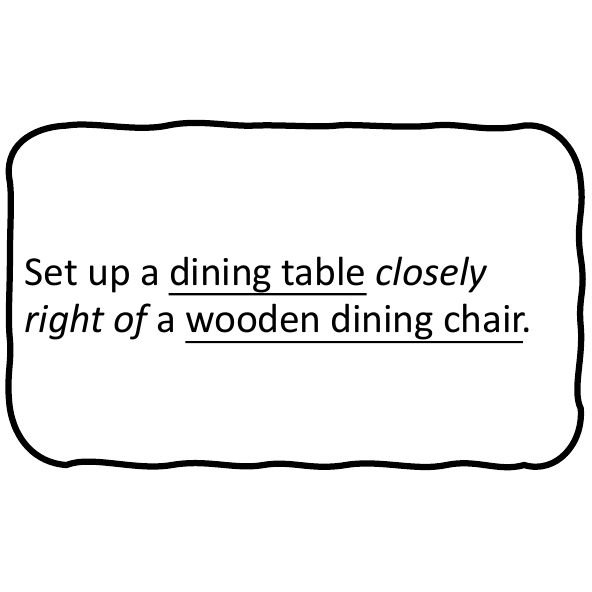}
  \caption{Text Prompt}
\end{subfigure}\hfill
\begin{subfigure}[t]{0.15\textwidth}
  \centering
  \includegraphics[width=\linewidth]{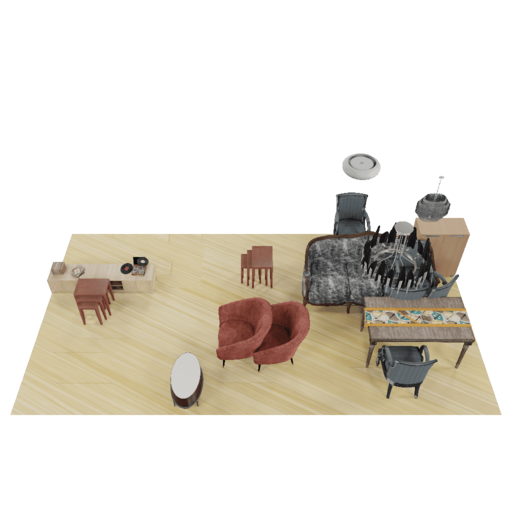}
  \caption{ATISS}
\end{subfigure}\hfill
\begin{subfigure}[t]{0.15\textwidth}
  \centering
  \includegraphics[width=\linewidth]{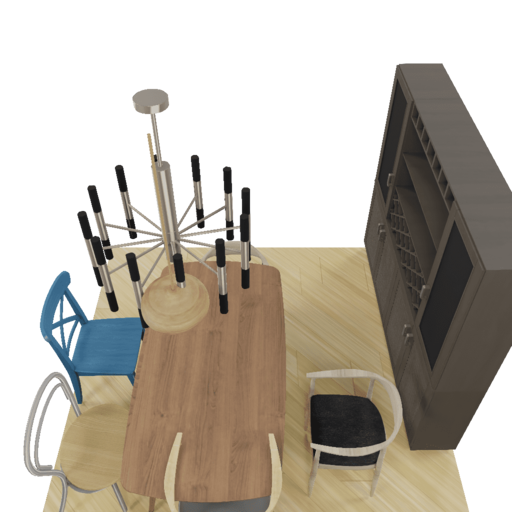}
  \caption{DiffuScene}
\end{subfigure}\hfill
\begin{subfigure}[t]{0.15\textwidth}
  \centering
  \includegraphics[width=\linewidth]{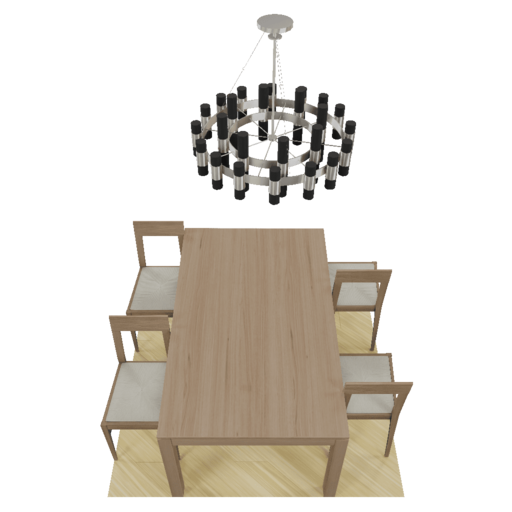}
  \caption{InstructScene}
\end{subfigure}\hfill
\begin{subfigure}[t]{0.15\textwidth}
  \centering
  \includegraphics[width=\linewidth]{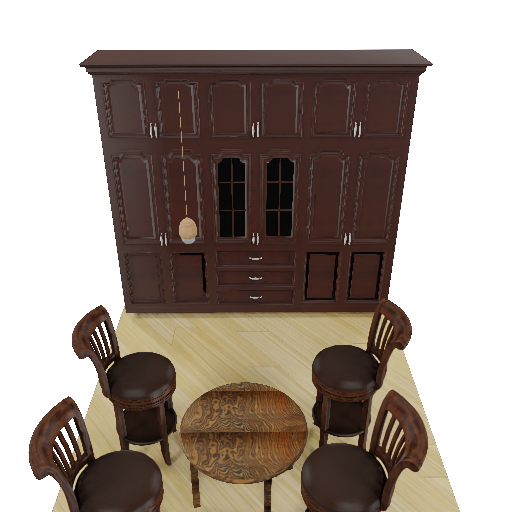}
  \caption{Our Method}
\end{subfigure}

\caption[Visual comparison.]{Visual comparison of our method and baseline approaches for text-guided scene generation on the living room (top row), and dining room (bottom row) datasets.}
\label{fig:visual-comparisons}
\end{figure*}

\begin{figure*}[h]
\centering
    \begin{subfigure}{0.15\textwidth}
    \includegraphics[width=\linewidth]{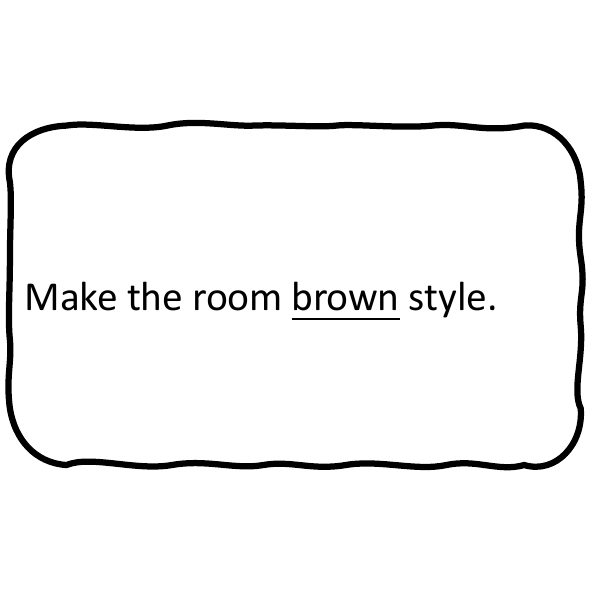}
    \caption{Prompt (Stylization)}
    \end{subfigure}
    \hfill
    \begin{subfigure}{0.15\textwidth}
        \includegraphics[width=\linewidth]{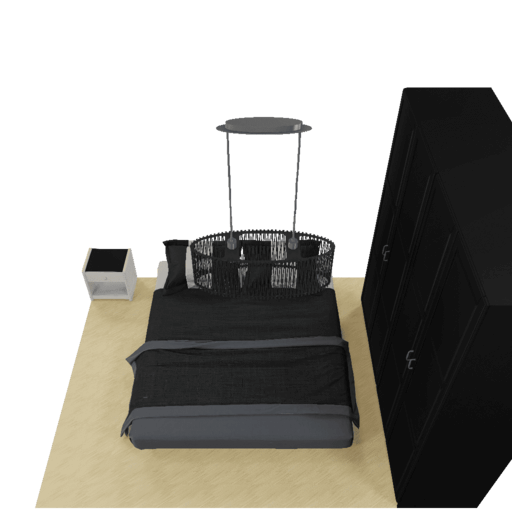}
    \caption{Original Scene}
    \end{subfigure}
    \hfill
    \begin{subfigure}{0.15\textwidth}
        \includegraphics[width=\linewidth]{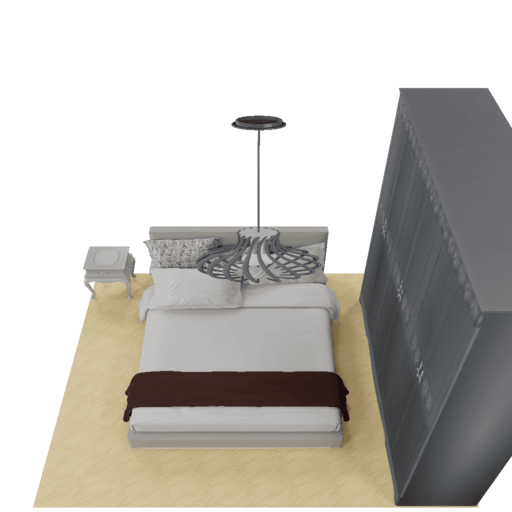}
    \caption{DiffuScene}
    \end{subfigure}
    \hfill
    \begin{subfigure}{0.15\textwidth}
        \includegraphics[width=\linewidth]{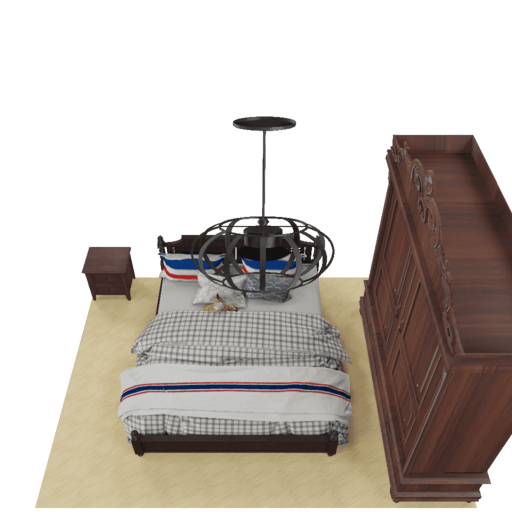}
        \caption{InstructScene}
    \end{subfigure}
    \hfill
    \begin{subfigure}{0.15\textwidth}
        \includegraphics[width=\linewidth]{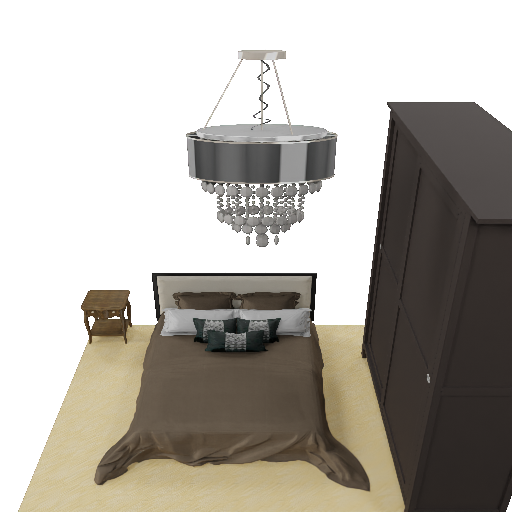}
    \caption{Our Method}
    \end{subfigure}
    \\ 
    \begin{subfigure}{0.15\textwidth}
        \includegraphics[width=\linewidth]{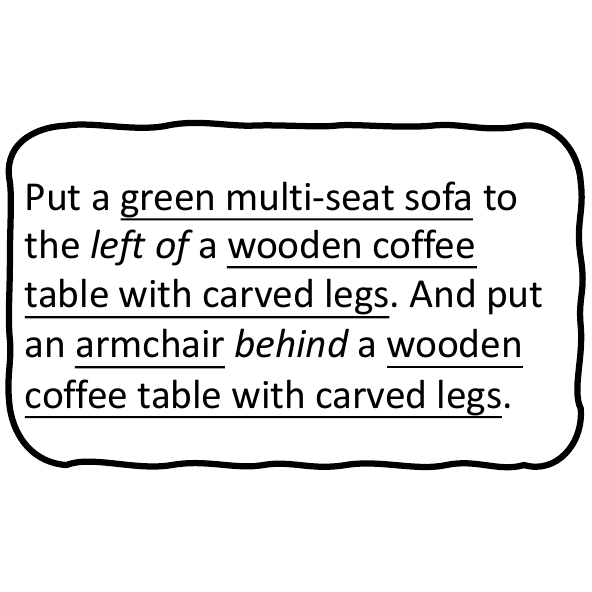}
        \caption{Prompt (Rearr.)}
    \end{subfigure}
    \hfill
    \begin{subfigure}{0.15\textwidth}
        \includegraphics[width=\linewidth]{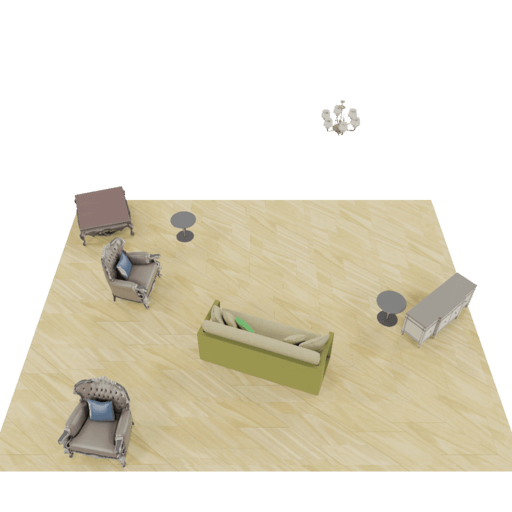}
        \caption{Messy Scene}
    \end{subfigure}
    \hfill
    \begin{subfigure}{0.15\textwidth}
        \includegraphics[width=\linewidth]{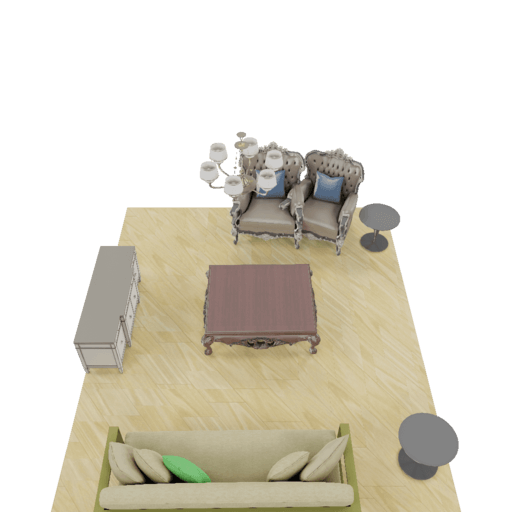}
        \caption{DiffuScene}
    \end{subfigure}
    \hfill
    \begin{subfigure}{0.15\textwidth}
        \includegraphics[width=\linewidth]{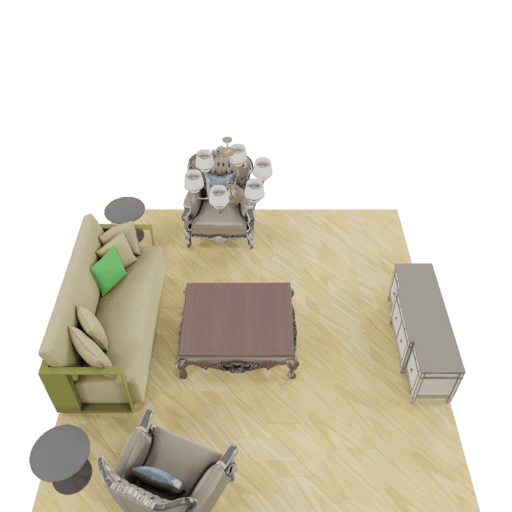}
        \caption{InstructScene}
    \end{subfigure}
    \hfill
    \begin{subfigure}{0.15\textwidth}
        \includegraphics[width=\linewidth]{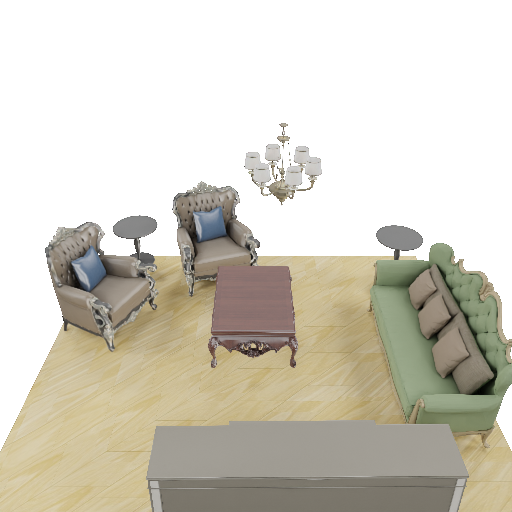}
        \caption{Our Method}
    \end{subfigure}
    \\
    \begin{subfigure}{0.15\textwidth}
        \includegraphics[width=\linewidth]{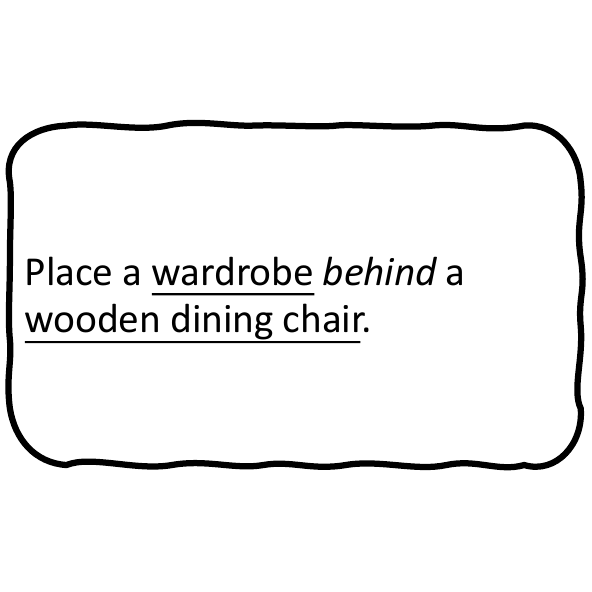}
    \caption{Prompt (Completion)}
    \end{subfigure}
    \hfill
    \begin{subfigure}{0.15\textwidth}
        \includegraphics[width=\linewidth]{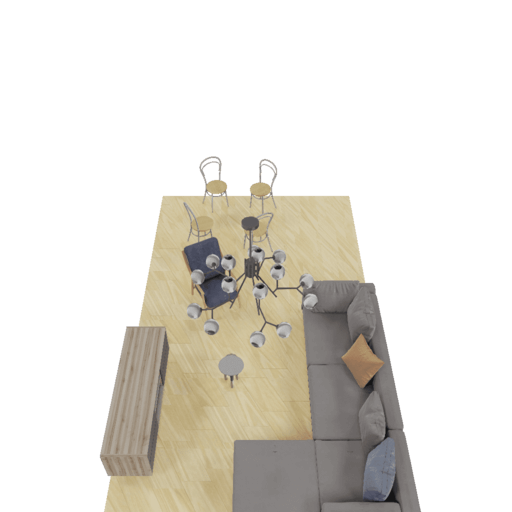}
    \caption{Partial Scene}
    \end{subfigure}
    \hfill
    \begin{subfigure}{0.15\textwidth}
        \includegraphics[width=\linewidth]{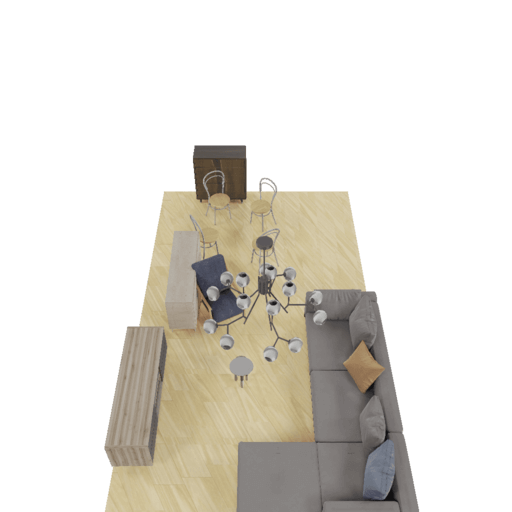}
    \caption{DiffuScene}
    \end{subfigure}
    \hfill
    \begin{subfigure}{0.15\textwidth}
        \includegraphics[width=\linewidth]{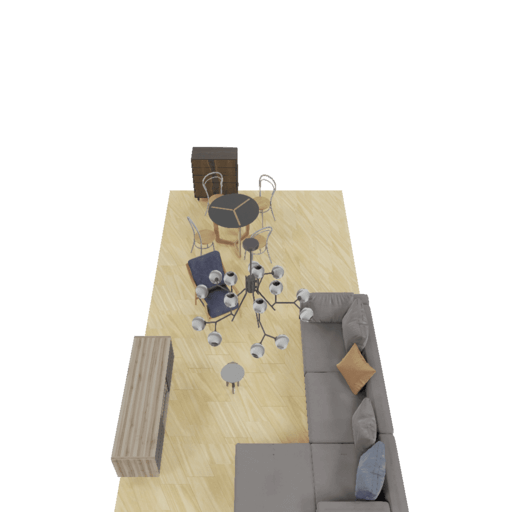}
    \caption{InstructScene}
    \end{subfigure}
    \hfill
    \begin{subfigure}{0.15\textwidth}
        \includegraphics[width=\linewidth]{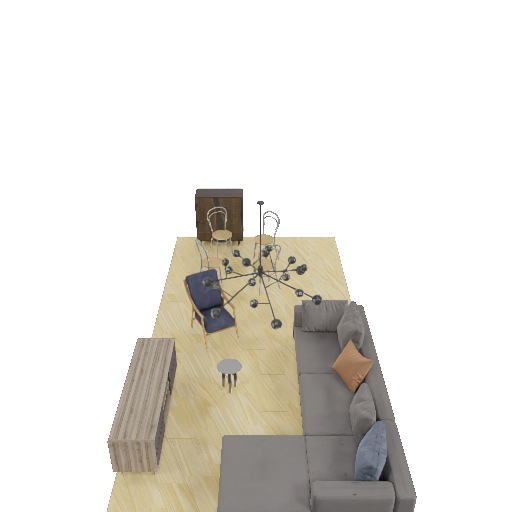}
    \caption{Our Method}
    \end{subfigure}
    \\
    \begin{subfigure}{0.20\textwidth}
        \includegraphics[width=\linewidth]{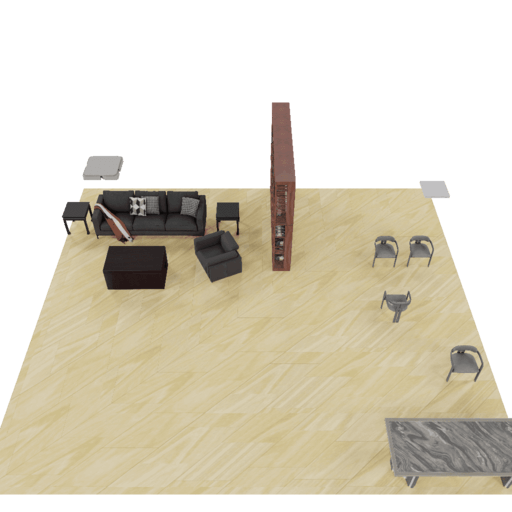}
        \caption{ATISS (Unconditional)}
    \end{subfigure}
    \hfill
    \begin{subfigure}{0.20\textwidth}
        \includegraphics[width=\linewidth]{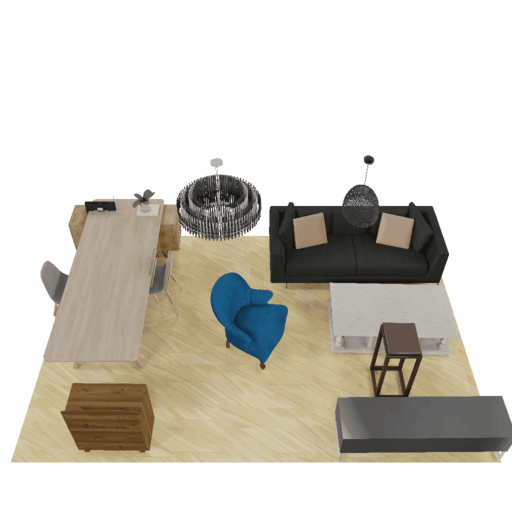}
        \caption{DiffuScene (Unconditional)}
    \end{subfigure}
    \hfill
    \begin{subfigure}{0.20\textwidth}
        \includegraphics[width=\linewidth]{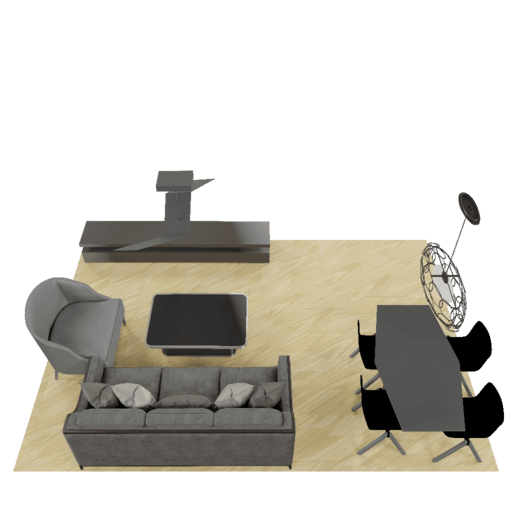}
         \caption{InstructScene (Unconditional)}
    \end{subfigure}
    \hfill
    \begin{subfigure}{0.20\textwidth}
        \includegraphics[width=\linewidth]{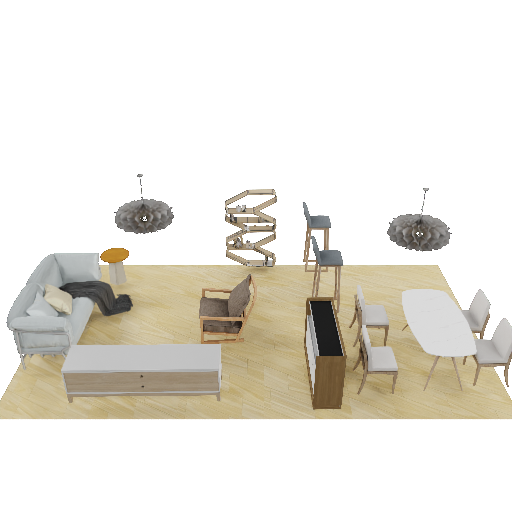}
        \caption{Our Method (Unconditional)}
    \end{subfigure}
\caption[Comparison visualizations.]{Comparison visualizations of our method with DiffuScene\cite{tang2024diffuscene} and InstructScene\cite{lin2024instructscene} for different zero-shot tasks. Due to space constraints, we omit ATISS \cite{paschalidou2021atiss} from the first three visualizations.}
    
    \label{fig:zero-shot}
\end{figure*}

\textbf{Datasets}. All experiments are carried out on the Scene Instruction 3D-FRONT dataset \footnote{The augmented version of 3D-FRONT from InsctructScene is available at \url{https://huggingface.co/datasets/chenguolin/InstructScene_dataset}.}, an augmented version of 3D-FRONT \cite{fu20213d}. 3D-FRONT is a professionally designed synthetic dataset, comprising 6,813 houses with 14,629 rooms, where each room is furnished with high-quality 3D furniture objects extracted from the 3D-FUTURE dataset \cite{fu20213d}. However, the original dataset lacks textual descriptions of room layouts and object appearances. To address this, InstructScene \cite{lin2024instructscene} augments it by extracting view-dependent spatial relations using predefined geometric rules (see Table 4 in InstructScene \cite{lin2024instructscene}). Then, 3D object captions are generated using BLIP \cite{li2022blip} and further refined with ChatGPT \cite{ouyang2022training}. The final scene instructions are created by randomly selecting relation triplets in the form \textit{(object, relation, subject)}, where both the \textit{object} and \textit{subject} are object names or object descriptions, and the \textit{relation} is chosen from a predefined set of spatial relationships.  For further details on dataset creation, we refer to Appendix A of InstructScene \cite{lin2024instructscene}. To ensure consistency with previous methods, we adopt the same filtering process and train/test split proposed by ATISS \cite{paschalidou2021atiss}. The filtering process removes duplicates and problematic scenes, resulting in 4041 bedrooms, 900 dining rooms and 813 living rooms.

\noindent
\textbf{Baselines}. We compare our method against three state-of-the-art task-specific methods for 3D indoor scene synthesis. (1) ATISS \cite{paschalidou2021atiss}, an autoregressive approach that employs a transformer encoder to predict object features. (2) DiffuScene \cite{tang2024diffuscene}, a diffusion-based method that utilizes a 1D U-Net to learn the denoising process of object features. (3) InstructScene \cite{lin2024instructscene}, a two-stage approach that first employs a discrete diffusion method to generate a scene graph prior, and then uses it to condition a continuous diffusion model to generate the final layout. These three approaches differ in their controllability. ATISS does not support text conditioning. Although DiffuScene accepts text prompts, it struggles with object descriptions because its shape codes were trained solely on point clouds without explicit text alignment. InstructScene addresses this by extending the first two approaches with a pretrained CLIP text encoder for text conditioning and applying the same feature quantizer to the OpenCLIP \texttt{ViT-bigG/14} embeddings. Building on this work, we compare our method against all three models (with extensions to (1) and (2) as described in InstructScene).

\noindent
\textbf{Evaluation Metrics}. 
To assess how well synthesized scenes follow user prompt instructions, we use the "instruction recall" (\textit{iRecall}) metric introduced by InstructScene \cite{lin2024instructscene}. It measures the proportion of predicted triplets \((o_{\text{pred}}, r_{\text{pred}}, s_{\text{pred}})\), derived from the geometric rules applied during dataset construction, that match the ground-truth triplets \((o_{\text{gt}}, r_{\text{gt}}, s_{\text{gt}})\) from the  text prompts. As in previous works \cite{paschalidou2021atiss,tang2024diffuscene,lin2024instructscene}, we evaluate scene quality by rendering \(256\times256\) top-view projections of both real and generated scenes, and compute standard metrics from image generation: Fréchet inception distance (FID) \cite{heusel2017gans}, FID-CLIP \cite{kynkaanniemi2022role}, kernel inception distance (KID) \cite{binkowski2018demystifying}, and scene classification accuracy (SCA). All images are rendered in Blender \cite{community2018blender}.

\begin{table*}[!t]
\centering
\footnotesize
\begin{tabular}{|c|c|c|c|c|c|c|}
\hline
Type of Room & Method & $\uparrow$ iRecall (\%) & $\downarrow$ FID & $\downarrow$ FID-CLIP & $\downarrow$ KID ($\times10^{-3}$) & SCA (\%) \\ 
\hline

\multirow{4}{*}{Bedroom} 
& ATISS\(^\dagger\)        & 48.13 & 119.73 & 6.95 & 0.39 & 59.17 \\  
& DiffuScene\(^\dagger\)   & \textbf{56.43} & 123.09 & 7.13 & 0.39 & 60.49 \\ 
& InstructScene           & 33.06 (66.53\(^*\)) & \textbf{119.32} & \textbf{6.66} & \textbf{0.30} & 60.49 \\ 
& GeoSceneGraph          & 30.61 & 121.21 & 6.91 & 0.92 & \textbf{53.70} \\ 
\hline

\multirow{4}{*}{Living room} 
& ATISS\(^\dagger\)        & 29.50 & 117.67 & 6.08 & 17.60 & 69.38 \\  
& DiffuScene\(^\dagger\)   & \textbf{31.15} & 122.20 & 6.10 & 16.49 & 72.92 \\ 
& InstructScene           & 28.91 (51.02\(^*\)) & \textbf{109.21} & 5.06 & 3.96 & \textbf{58.33} \\ 
& GeoSceneGraph          & 25.17 & 111.20 & \textbf{5.01} & \textbf{3.81} & 59.37 \\ 
\hline

\multirow{4}{*}{Dining room} 
& ATISS\(^\dagger\)        & 37.58 & 137.10 & 8.49 & 23.60 & 67.61 \\  
& DiffuScene\(^\dagger\)   & 37.87 & 145.48 & 8.63 & 24.08 & 70.57 \\ 
& InstructScene           & 37.92 (58.74\(^*\)) & \textbf{122.22} & \textbf{6.20} & \textbf{0.41} & \textbf{55.68} \\ 
& GeoSceneGraph          & \textbf{40.89} & 123.04 & 6.41 & 0.49 & 56.81 \\ 
\hline
\end{tabular}
\caption[Quantitative comparisons of scene synthesis guided by text prompts.]{Quantitative comparisons of scene synthesis guided by text prompts. The \(\dagger\) symbol indicates that these methods could not be reproduced, and their metrics were taken from the InstructScene \cite{lin2024instructscene} paper. Results marked with \(^*\) in parentheses denote the \textit{iRecall} computed on the semantic scene graph of the first stage of InstructScene. For the SCA metric, results close to 50\% indicate better performance.}

\label{tab:main-table}
\end{table*}
\begin{table*}[!t]
\centering
\footnotesize
\begin{tabular}{|c|c|c|c c|c c|c|}
\hline
\multirow{2}{*}{Type of Room} & \multirow{2}{*}{Method} & Stylization & \multicolumn{2}{c|}{Re-arrangement} & \multicolumn{2}{c|}{Completion} & Uncond. \\ 
                             &  & $\uparrow \Delta \times 1e^{-3} $ & $\uparrow$ iRecall  & $\downarrow$ FID & $\uparrow$ iRecall  & $\downarrow$ FID & $\downarrow$ FID \\ \hline
\multirow{3}{*}{Bedroom} & ATISS\(^\dagger\) & \textbf{3.44} & 61.22 & 107.67 & 64.90 & 89.77 &  134.51 \\ 
                         & DiffuScene\(^\dagger\) & 1.08 & \textbf{68.57} & \textbf{106.15} & 48.57 & 96.28 & 135.46 \\ 
                         & InstructScene  & 1.59 & 38.37 & 113.96 & \textbf{65.71} & 84.41 & \textbf{124.97}\\ 
                         & GeoSceneGraph & 0.5 & 41.63 & 120.01 & 46.53 & \textbf{84.14} & 129.52 \\ \hline
\multirow{3}{*}{Living room} & ATISS\(^\dagger\) & -3.57 & 31.97 & 117.97 & 43.20 & 106.48 & 129.23 \\ 
                             & DiffuScene\(^\dagger\) & -2.69 & \textbf{41.50} & 115.30 & 19.73 & 95.94 & 129.75 \\ 
                             & InstructScene  & \textbf{-0.95} & 37.76 & \textbf{109.46} & \textbf{48.64} & 81.39 & \textbf{111.34}\\ 
                             & GeoSceneGraph   & -1.94 & 39.46 & 133.67 & 20.75 & \textbf{81.09} & 116.74 \\ \hline
\multirow{3}{*}{Dining room} & ATISS\(^\dagger\) & -1.11 & 36.06 & 134.54 & \textbf{57.99} & 122.44 & 147.52 \\ 
                             & DiffuScene\(^\dagger\) & -2.98 & 46.84 & 133.73 & 32.34 & 115.08 & 150.81 \\ 
                             & InstructScene  & \textbf{-1.75} & 46.84 & \textbf{123.95} & 56.13 & 96.13 & \textbf{126.97}\\ 
                             & GeoSceneGraph   & -2.22 & \textbf{47.96} & 129.02 & 35.69 & \textbf{83.19} & 134.90 \\ \hline
\end{tabular}
\caption[Quantitative comparisons on zero-shot applications.]{Quantitative comparisons on four zero-shot applications for scene synthesis. Symbol \(\dagger\) indicates that these methods could not be reproduced, and their metrics were taken from the InstructScene \cite{lin2024instructscene} paper.}
\label{tab:zero-shot}
\end{table*}
\begin{table*}[!t]
\centering
\footnotesize
\label{tab:ablation}
\begin{tabular}{|c|c c|c|c|c|c|}
\hline
 \multirow{2}{*}{Method} & \multicolumn{2}{c|} {$\uparrow$ iRecall (\%)} & \multirow{2}{*}{$\downarrow$FID} & \multirow{2}{*}{$\downarrow$FID-CLIP} & \multirow{2}{*}{$\downarrow$ KID ($\times1e^{-3}$)} &\multirow{2}{*}{SCA (\%)}  \\ 
& Standard & Rot. Inv. & & & & \\
\hline
 Our Method  & 25.17 & \textbf{59.52} & \textbf{111.20} & \textbf{5.01} & \textbf{3.81} & \textbf{59.37} \\  
Concat. Conditioning \cite{hoogeboom2022equivariant}& 16.33 & 27.21 & 131.04 & 7.55 & 9.62 & 63.42 \\
Cross Attention  & \textbf{27.21} & 52.38 & 115.88 & 5.40 & 4.48 & 59.37 \\ 
\hline
\end{tabular}
\caption[Quantitative comparisons for the ablation study.]{Results for our ablation studies on the living room dataset. Concat. Conditioning refers to concatenate the text and time step embeddings to the node features. Cross Attention refers to using a cross attention module with the node features and the text embeddings.}
\label{tab:ablation}
\end{table*}

\subsection{Text-guided 3D Scene Synthesis}
\label{sec:text-guided-3D-scene-synthesis}

In Tables~\ref{tab:main-table} and \ref{tab:zero-shot}, we present the experimental results for both InstructScene and our method. Note that there are discrepancies between our reproduced metrics and those reported in the InstructScene paper \cite{lin2024instructscene}. Based on our review of their code, it appears that the reported metrics in the InstructScene paper were computed on the output of the first stage (i.e., the semantic scene graph) rather than the final layout. Since this metric can only be computed for their method, we also include these results in parentheses for comparison. Because the official extensions of ATISS and DiffuScene are not publicly available, we were unable to reproduce them and instead report their results directly from the InstructScene paper. Table~\ref{tab:main-table} provides a quantitative comparison of our method and the baselines on the task of 3D scene synthesis guided by text instructions.

We first evaluate scene quality using FID, FID-CLIP, KID, and SCA. The results show that methods incorporating a scene graph into their pipeline, specifically InstructScene and our GeoScene- 
Graph, tend to produce higher-quality scenes. This improvement is particularly evident in more complex environments such as living and dining rooms, which can contain up to 21 objects and more than 10 objects on average. In contrast, the bedroom dataset is comparatively simpler, containing at most 12 objects and typically around 5 per scene. Although InstructScene reports slightly stronger performance, our method achieves consistently competitive results across all datasets. For controllability, measured by the \textit{iRecall} metric, the results are more varied. Our method outperforms all others on the dining room dataset and remains competitive on the living room dataset, but is surpassed on the bedroom dataset, where DiffuScene and ATISS achieve substantially higher scores. We hypothesize two contributing factors: (1) the bedroom dataset exhibits lower layout diversity, with only about five objects per scene (on average) and many layouts being structurally similar, making it easier for methods that do not incorporate scene graph structure to memorize these patterns. In such cases, the inherent scene graph becomes less informative, which may advantage these methods since they do not rely on the graph structure of the 3D scene; and (2) there may have been modifications to the evaluation procedure in the InstructScene implementation to enable comparisons using semantic scene graphs, which could have influenced the reported results.

Figure~\ref{fig:visual-comparisons} presents qualitative comparisons between our method and the baselines on the living room, and dining room datasets.

\subsection{Zero-shot Applications}
\label{sec:zero-shot-applications}
In addition to the core task discussed in Section \ref{sec:text-guided-3D-scene-synthesis}, we evaluate our method on four zero-shot applications proposed in InstructScene \cite{lin2024instructscene}: stylization, re-arrangement, completion, and unconditional generation. Stylization modifies an input scene based on a text prompt specifying a style change. Re-arrangement repositions misplaced objects to match the text description. Completion augments a partial scene with additional objects that match the text prompt. Conversely, unconditional generation produces scenes with no text control or prior scene state. For stylization, we use the difference cosine similarity \(\Delta\), proposed in InstructScene, to evaluate how well the style matches the text prompt. 
For a scene \(S_k\) with \(N_k\) objects, it is defined as:
\begin{equation}
    \Delta_{k} = \frac{1}{N_k}\sum_{i=0}^{N_k} S_C(e_{ki}^{obj}, e_{ki}^{style}) - S_C(e_{ki}^{obj}, e_{ki}^{class}),  
\end{equation}%
where \(S_C\) is the cosine similarity, \(e_{ki}^{obj}\) are CLIP visual features of 
object \(o_i\), \(e_{ki}^{style}\) are CLIP text features from the stylization prompt, 
and \(e_{ki}^{class}\) are text features of the object class. 
The final score is the average across all \(M\) scenes: \(\Delta = \frac{1}{M}\sum_{k=1}^{M}\Delta_k\). For the re-arrangement and completion tasks, we report both \textit{iRecall} and FID metrics, while for unconditional generation, we report only the FID metric as it is not condition on anything.
Although our method does not top the zero-shot tasks, it shows competitive performance in controllability and generation quality. Notably, for unconditional generation, it ranks second, outperforming ATISS and DiffuScene by a wide margin. Figure~\ref{fig:zero-shot} shows qualitative results on the zero-shot tasks. The visualizations indicate that our method performs strongly, especially in rearrangement and unconditional generation scenarios.

\subsection{Ablation Studies}
\label{sec:ablation}

We focus our ablation studies on two conditioning strategies for the EGNN. The first strategy concatenates the text and diffusion time-step embeddings with the node features before processing them through the EGCL layers, as implemented in the original EDM \cite{hoogeboom2022equivariant}. The second strategy employs a cross-attention instead of a self-attention module (see Figure \ref{fig:egnn-architecture}) that fusions node features with text embeddings. For this ablation study, we also introduce the \textit{iRecall (rotation-invariant)} metric to enable a more comprehensive evaluation. This metric evaluates multiple rotated versions of the scene, addressing the view-dependence of the original \textit{iRecall} metric. The \textit{iRecall rotation invariant} metric computes \textit{iRecall} on four rotated versions of a generated scene (0°, 90°, 180°, 270°) and takes the maximum:
\begin{equation}
\ \textit{iRecall}_{\textit{RI}} 
    = \max_{\theta \in \{0,90,180,270\}}\bigl[\textit{iRecall}(R_\theta(S))\bigr],
\end{equation}
where \(R_\theta\) is the rotation matrix for angle \(\theta\) and \(S\) is the 
generated scene. 

Table~\ref{tab:ablation} reports the experimental results on the living room dataset. The results confirm that simply concatenating the time and text context to the node features is insufficient. Furthermore, even a more sophisticated cross-attention strategy performs worse once the view-dependence is removed. In contrast, our approach, which first incorporates the time condition using a ResNet and self-attention Transformer blocks, and then injects the text context at each message-passing step of the EGNN, achieves superior performance.

\section{Conclusion}
\label{sec:conclusion}
In this work, we presented GeoSceneGraph, a diffusion-based approach for synthesizing 3D indoor scenes from text prompts. Our method leverages the inherent geometric graph structure of the scene and does not rely on ground-truth semantic edges for object relationships, enabling edge-level open-vocabulary training. Despite not using semantic edge annotations, our method achieves competitive results in both visual quality and text controllability, as demonstrated in our experiments. To handle high-dimensional conditioning for the equivariant graph neural network (EGNN), we introduce a strategy that first incorporates the time condition using a ResNet and self-attention Transformer blocks, and then integrates the text context within the message-passing steps of the EGNN. We corroborate these findings in our ablation studies.

{
	\begin{spacing}{1.17}
		\normalsize
		\bibliography{main} %
	\end{spacing}
}

\end{document}